\documentclass{article} 
\usepackage{iclr2024_conference,times}

\usepackage{enumitem}

\usepackage{hyperref}
\usepackage{url}
\usepackage{graphicx}
\usepackage{subcaption}
\usepackage{tabularx}
\usepackage{booktabs}
\usepackage{algorithm2e}
\usepackage{wrapfig}
\usepackage{amsmath}
\usepackage{amssymb}
\usepackage[most]{tcolorbox}
\newtcolorbox{mycodebox}[2][]{
  breakable,
  title=#2, 
  colback=gray!5,
  colframe=gray!80,
  colbacktitle=black!70, 
  coltitle=white, 
  fonttitle=\bfseries, 
  left=10pt,
  right=10pt,
  top=10pt,
  bottom=10pt,
  boxsep=0pt,
  arc=4mm, 
  outer arc=4mm, 
  toptitle=2mm, 
  bottomtitle=2mm, 
  #1 
}

\title{SeedEdit:\\Align Image Re-Generation to Image Editing}

\newcommand{\sharedthanks}{\thanks{Equal Contribution}}

\author{
Yichun Shi\footnotemark[1], ~Peng Wang\sharedthanks, ~Weilin Huang \\
$~$Seed Team, $~$ByteDance \\
}

\iclrfinalcopy 
\begin{document}

\maketitle

\begin{abstract}
We introduce SeedEdit, a diffusion model that is able to revise a given image with any text prompts. In our perspective, the key to such a task is to obtain an optimal balance between maintaining the original image, i.e. \textit{image reconstruction}, and generating a new image, i.e. \textit{image re-generation}. To this end, we start from a weak generator (text-to-image model) that creates diverse pairs between such two directions and gradually align it into a strong image editor that well balances between the two tasks. SeedEdit can achieve more diverse and stable editing capability over prior image editing methods, enabling sequential revision over images generated by diffusion models. Our website is \url{https://team.doubao.com/seededit}.
\end{abstract}

\begin{figure}[t!]
    \includegraphics[width=1.05\linewidth]{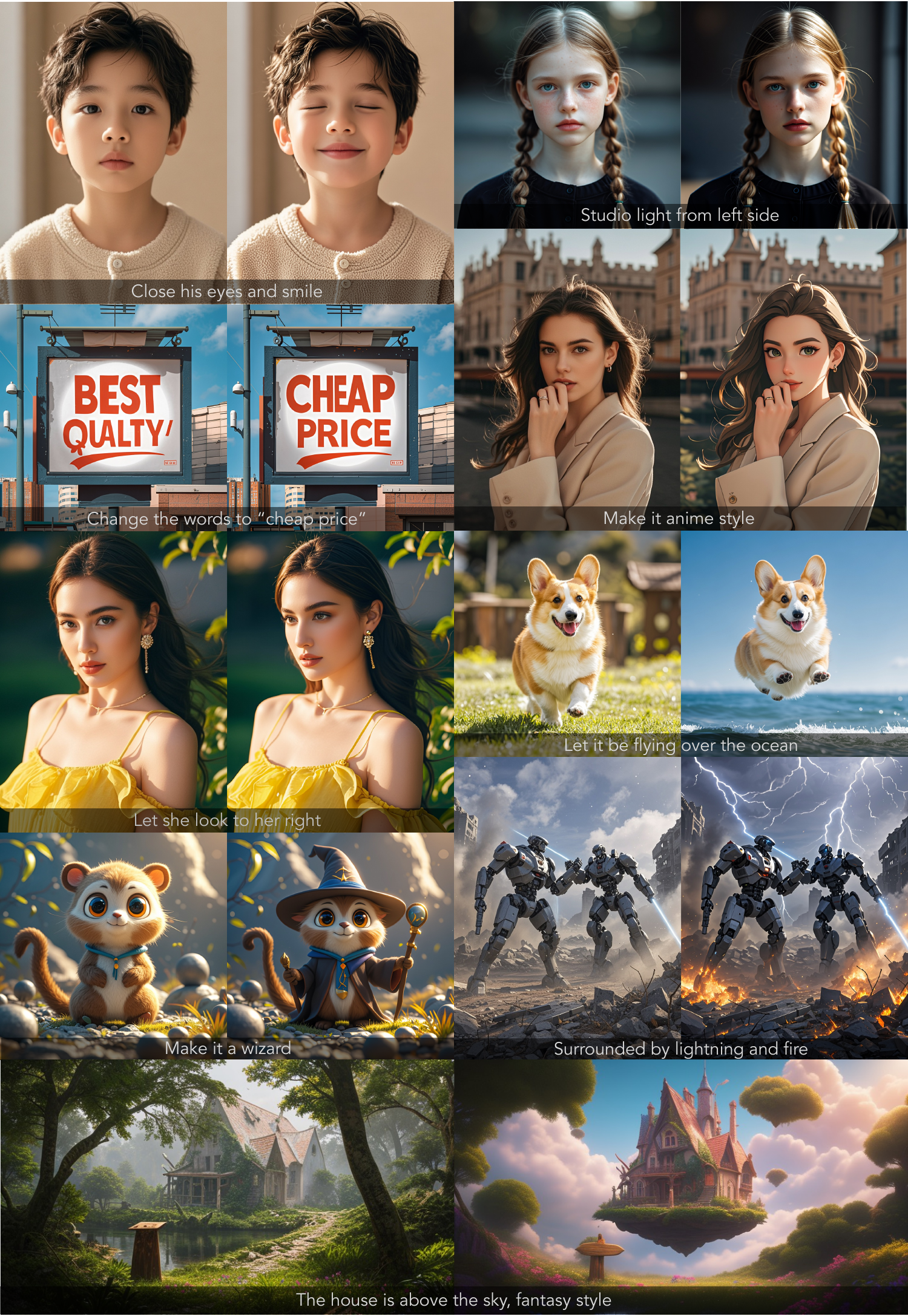}
    \caption{Example images edited by our method with one unified model and instructions only.}
    \label{fig:teaser}
\end{figure}

\section{Introduction}

Today's diffusion models can create realistic and diverse images from only text descriptions. However, these generated images are usually quite uncontrollable and to some extent, the generation process is like throwing a dice until one sees a good output. 
To obtain more controllability over the generated content, a desired feature is instructional image editing, \textit{i.e.} revising an input image with text descriptions. This can be regarded as a intersection between image generation and image understanding, both of which are quite mature today. Yet to this date, the technology of image editing itself still falls far behind both generation and understanding. 


Existing image editing for diffusion models can be roughly categorized into two types. Firstly, the training-free methods combine specific techniques such as DDIM Inversion~\citep{nichol2021glide,mokady2023null}, test-time fine-tuning~\citep{ruiz2023dreambooth,kawar2023imagic}, attention control~\citep{cao2023masactrl,hertz2022prompt2prompt} to reconstruct an input image and generate a new one with the new text guidance. But since both the reconstruction and the re-generation process suffer from instability, the combination of these two accumulates into more error into the edited image, which could be inconsistent with either the input image or the target description. 

The second type of methods are data driven approaches, where a large-scale pairwise editing dataset is prepared to train a instructional diffusion model~\citep{brooks2023instructpix2pix,zhang2024magicbrush,HQEdit,PIPE,zhao2024ultraedit}. The main difficulty here, however, is to prepare a diverse and high-quality editing dataset. Unlike image datasets that can be massively collected from the Internet, image editing pairs are very rare and it is almost impossible to collect a high-quality dataset that covers all types of editing pairs. So existing works attempt to use certain tools, such as Prompt-to-Prompt~\citep{hertz2022prompt2prompt} or in-painting to create such a dataset. But consequently, their performance is limited by these data creation tools, who themselves are not satisfying either.

To overcome the above mentioned difficulties, we introduce a new framework to convert an image generation diffusion model to one that edits images. We recognize that image editing is essentially a balance between image reconstruction and re-generation, and hence we develop a pipeline that first generates diverse pairwise data that scatters into these two directions, and then gradually align a image-conditioned diffusion model to arrive at an optimal balance between these two tasks. Overall, it leads to a model that is capable of revising images with either instructions or descriptions, which we call SeedEdit, and yields superior performance compared to prior studies.

\section{SeedEdit}

\begin{figure}[t]
\vspace{-0.1em}
\centering
\begin{subfigure}[b]{0.43\textwidth}
\includegraphics[width=\linewidth]{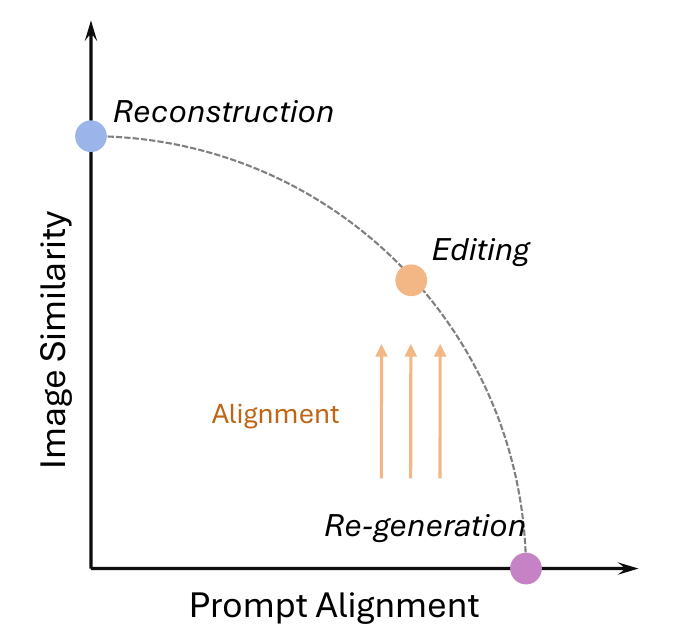}\hfill
\caption{Objective Illustration}
\end{subfigure}\hfill
\begin{subfigure}[b]{0.55\textwidth}
\includegraphics[width=\linewidth]{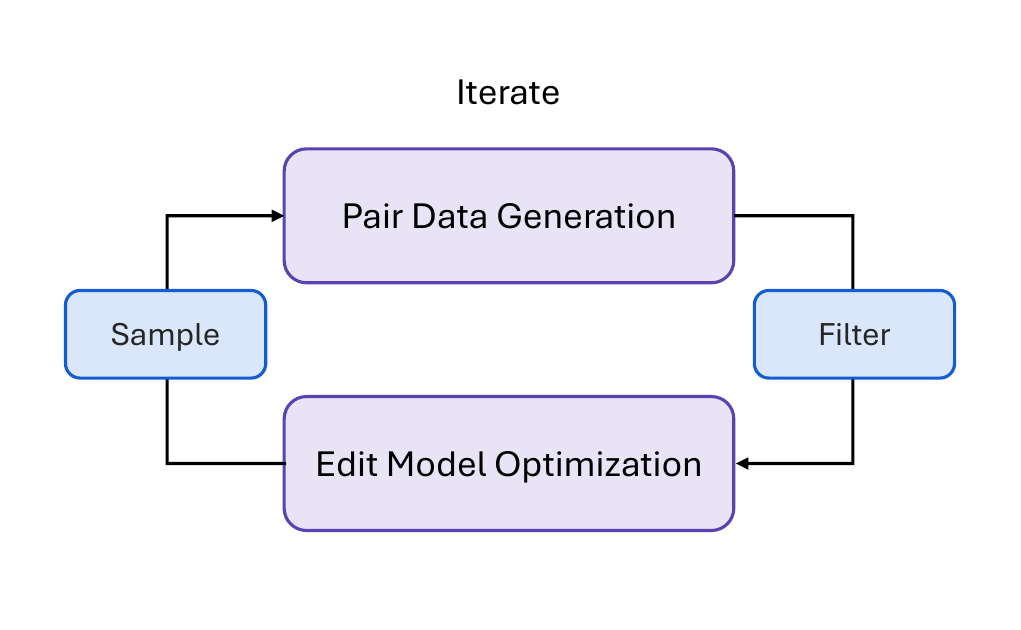}
\caption{Optimization}
\end{subfigure}
\vspace{-0.5em}\caption{Overview of SeedEdit framework. We align a T2I model as lower bound for editing by improving image consistency. 
\textbf{Right}: Our optimization pipeline, we have an init edit model based on T2I and then iteratively conduct data sampling and model optimization to reach the optimal balance.
}
\label{fig:workflow}
\end{figure}

The core difficulty of the image editing problem is the scarcity of pairwise image data. We address this problem from an alignment perspective. In particular, we regard text-to-image (T2I) model as a weak editing model, which achieves "editing" by generating a new image with a new prompt. We then distill and align such a weak editing model into a strong one by maximally inherit the re-generation capability while improving image consistency, as shown in Figure~\ref{fig:workflow}.

\subsection{T2I model for Editing Data Generation}
\label{sec:method:initial_data}


\begin{wrapfigure}{r}{0.48\textwidth}
\captionsetup{font=small}
\vspace{-1.8em}
\includegraphics[width=\linewidth]{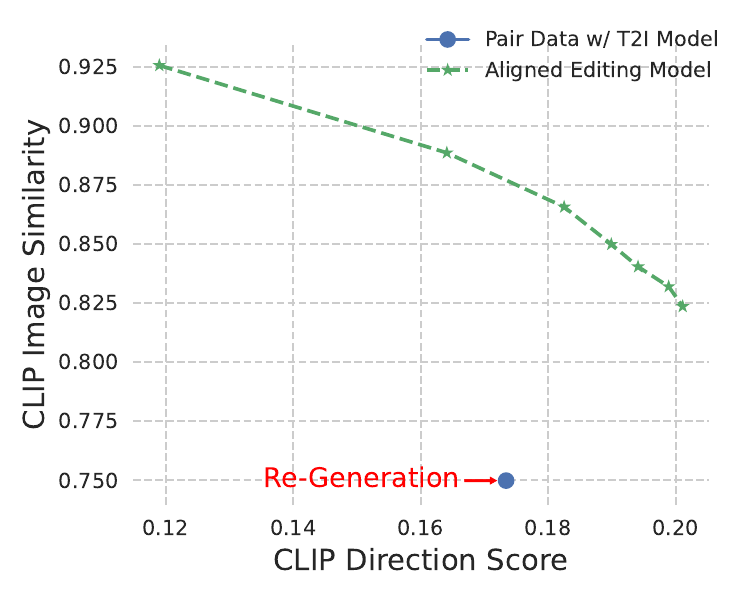}
\vspace{-2.5em}
\caption{
Our aligned editing model can achieve a similar or higher direction score (prompt alignment) with much higher image similarity compared to re-generation. The green curve is drawn by sampling different CFG for editing model.
}
\vspace{-0.8em}
\label{fig:t2i_vs_editing}
\end{wrapfigure}
Our initial editing data are generated using a pre-trained T2I model as an editing model, where a pair of images before and after editing can be generated with corresponding text descriptions, similar to InstuctPix2Pix~\cite{brooks2023instructpix2pix}. With such data, we could distill a T2I model into an image-conditioned editing model. However, such naive re-generation could lead to inconsistency between the two images. To improve consistency, there exist various approaches, such as prompt-to-prompt~\citep{hertz2022prompt2prompt,brooks2023instructpix2pix} and attention control~\cite{cao2023masactrl}. However, these techniques can generate very limited types of pair data and can hardly cover all types of image editing. Therefore, we combine different re-generation techniques and parameters to create a much more diverse dataset. In particular, we generate a large-scale pairwise dataset with more randomness to ensure diversity, and then we apply filters to choose good examples for model training and alignment. Fig~\ref{fig:t2i_vs_editing} illustrates that our aligned model performs much better than naive re-generation based on the CLIP metrics.

\subsection{Causal Diffusion Model with Image Input}
\label{sec:method:model_and_training}

\begin{figure}[t]
    \centering
    \begin{subfigure}[b]{0.49\textwidth}
        \centering
        \includegraphics[width=\linewidth]{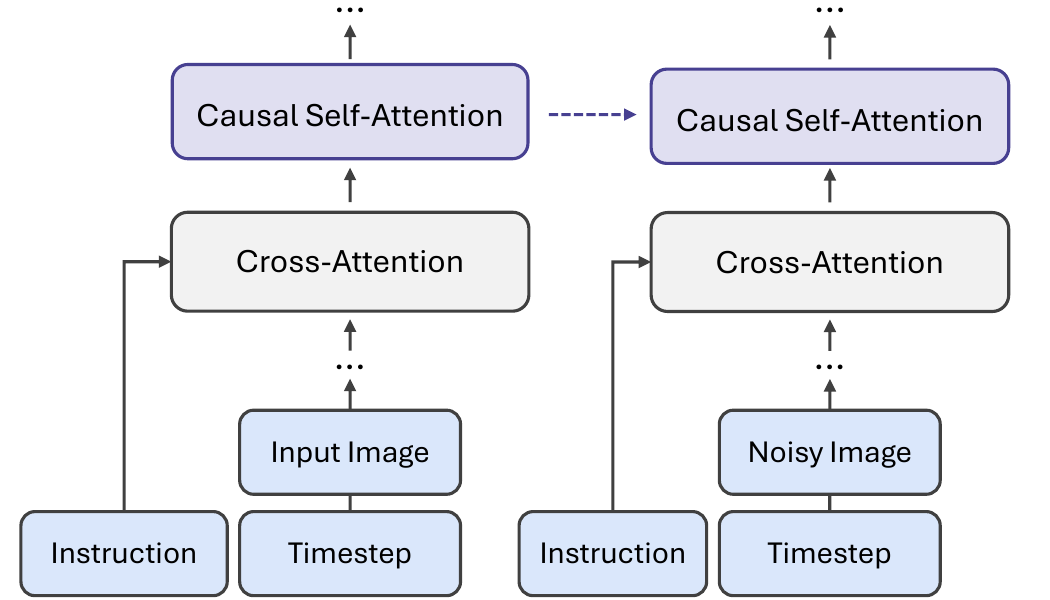}
        \caption{Causal UNet}
    \end{subfigure}\hfill
    \begin{subfigure}[b]{0.49\textwidth}
        \centering
        \includegraphics[width=\linewidth]{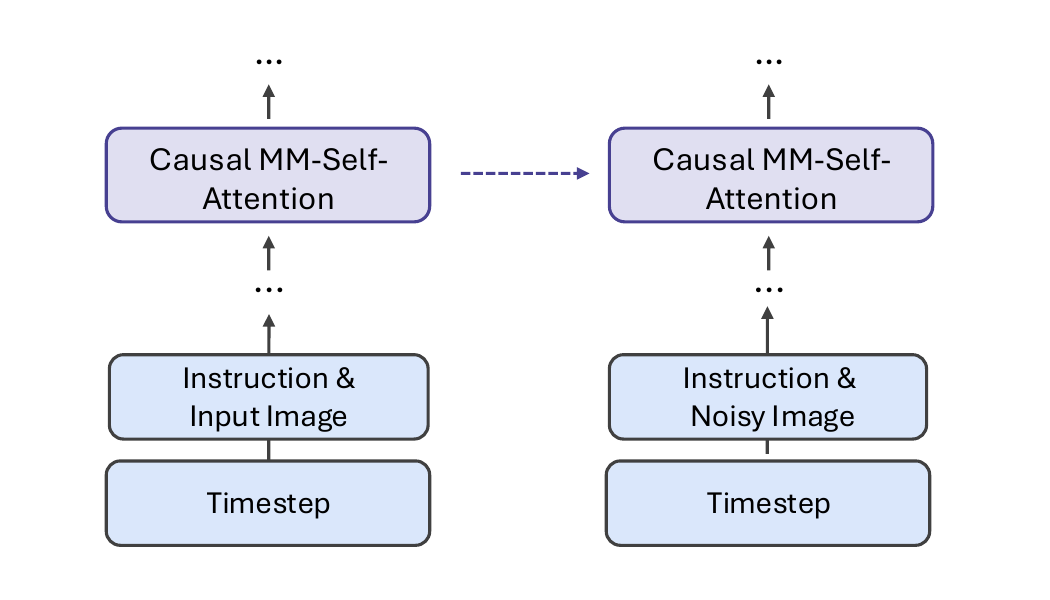}
        \caption{Causal MM-DiT}
    \end{subfigure}
    \caption{Architecture of causal diffusion model for image conditioning. Two branches with shared parameters are applied to the input image and instructions, respectively.}
    \label{fig:architecture}
\end{figure}

The model architecture of our image-conditioned diffusion model is shown in Fig.~\ref{fig:architecture}. Unlike previous studies that add additional input channels for image conditioning~\citep{brooks2023instructpix2pix}, we reuse self-attention for this purpose, where two branches of the diffusion model (shared parameters) are applied to the input and output image, respectively. This is inspired by prior training-free methods~\citep{cao2023masactrl} and we empirically found that such an architecture performs better on geometric deformation tasks and introduces fewer new parameters. Specifically, a causal self-attention structure is introduced such that two networks can build communications based on intermediate features. 
If we drop the input branch, it leads to the original T2I diffusion model, allowing for a mixed training on editing and T2I data.


\subsection{Iterative Alignment}
Because of the noisy dataset, the initial editing model trained on the pair of examples may not be sufficiently robust for applications. That is, like the dataset itself, the model is able to cover diverse editing tasks, but only with a limited success rate. To further ensure the robustness of the model, we propose progressively aligning the editing model by adding additional rounds of fine-tuning. In particular, since we already have an editing model at this stage, we may prepare a new set of data based on the current model following a similar pipeline for data generation. 
The results are then labeled and filtered again to fine-tune the editing model as in Sec.~\ref{sec:method:model_and_training}. We repeat this process for multiple rounds until the model converges, i.e. no more improvements over our metrics.

\section{Experiments}

\subsection{Benchmark and Metrics}
\label{sec:exp:benchmark_metrics}
Two base models are evaluated for our experiments, namely SDXL~\citep{SDXL} and an in-house T2I model based on DiT architecture~\citep{peebles2023DiT,SD3}.

We use two public datasets to evaluate image editing performance. The HQ-Edit dataset proposed in~\citep{HQEdit} and Emu Edit dataset from ~\citep{sheynin2024emu}. The former is composed of $293$ Dalle3 generated images and the latter is composed of $535$ real in-the-wild image inputs. We note that our method is mainly focused on the application scenarios in the HQ-Edit benchmark, where we want to revise T2I generated images with arbitrary instructions. Emu Edit is rather different from our training data, which mostly includes local editing on real-scene images. Therefore, we consider Emu Edit as an Out-of-Domain (OOD) test but mainly rely on HQ-Edit to evaluate the application potential of our method. 

We adopt two metrics to evaluate the editing performance. The first is CLIP-based~\citep{brooks2023instructpix2pix}, where CLIP Direction Score is used to evaluate the alignment of the editing prompt and the CLIP image similarity is used to measure consistency. The second is LLM-as-evaluator, where GPT is used to replace the CLIP Direction score to measure the success of the editing. 

\subsection{Image Editing Comparison}
We compare our method with several state-of-the-art image editing methods, including a training-free method Prompt-to-Prompt (Null-text Inversion)~\citep{hertz2022prompt2prompt,mokady2023null}, and data-driven methods Instruct-Pix2Pix~\citep{brooks2023instructpix2pix}, MagicBrush~\cite{zhang2024magicbrush}, Emu Edit~\cite{sheynin2024emu} and UltraEdit~\cite{zhao2024ultraedit}. Since Emu Edit is not open-sourced, we only compare them on their own test set. For the other methods, we used their model released with default parameters for comparison.
Table~\ref{tab:compare_baseline} shows the quantitative results of the baselines and our method. Overall, our method shows a significantly higher editing score on both benchmarks than open-source baselines. Meanwhile, we also observe a higher CLIP image similarity on the HQ-Edit dataset, which indicates a better preservation of the content in the original image. 

Although we mainly focus on the application scenario for revising T2I images as in HQ-Edit, our method also achieves descent quantitative scores on the Emu Edit benchmark, which is comparable/better to the original Emu Edit method. However, in general, we observe that the quality of the generated images of all methods (including ours) is not so satisfying on the Emu Edit benchmark, which proves our belief that the revisement of T2I images could be a first step to be solved before editing on arbitrary in-the-wild images.

\begin{figure}[t]
    \newcolumntype{Y}{>{\centering\arraybackslash}X}
    \begin{tabularx}{\linewidth}{YYYYY}
    \small Input & \small  Instruct-Pix2Pix & \small MagicBrush & \small UltraEdit & \small  SeedEdit (SDXL)\\
    \end{tabularx}
    \begin{subfigure}[b]{\textwidth}
        \centering
        \includegraphics[width=0.195\linewidth]{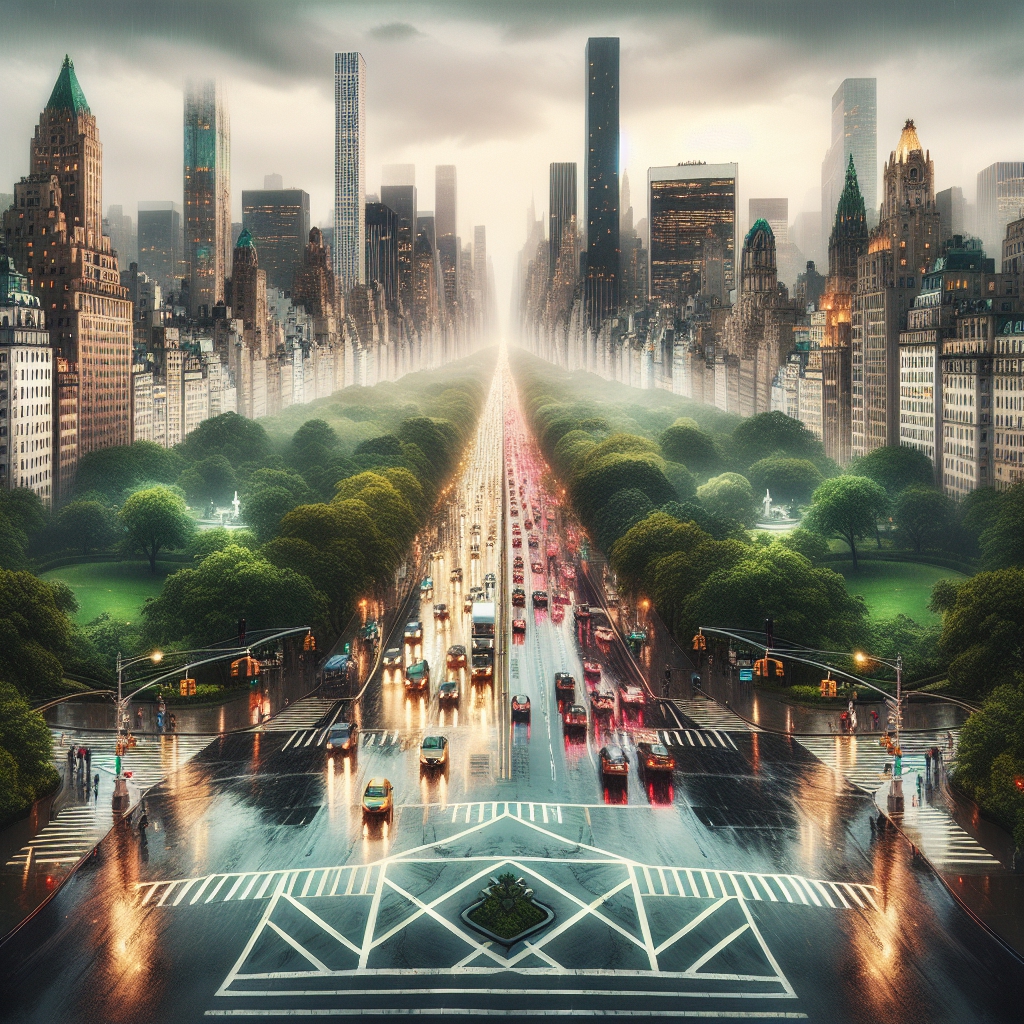}\hfill
        \includegraphics[width=0.195\linewidth]{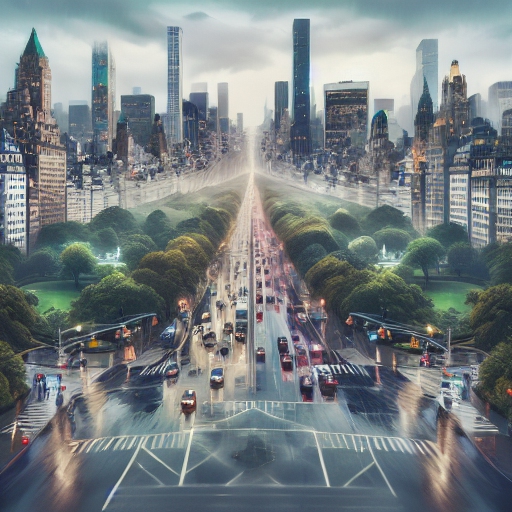}\hfill
        \includegraphics[width=0.195\linewidth]{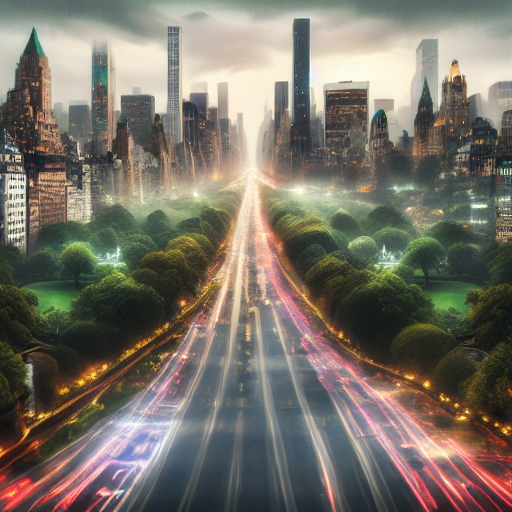}\hfill
        \includegraphics[width=0.195\linewidth]{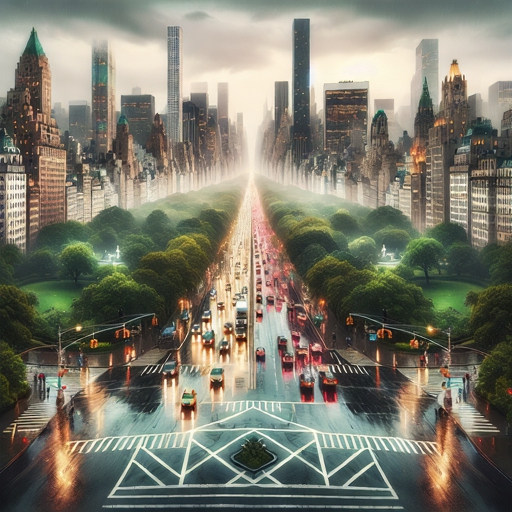}\hfill
        \includegraphics[width=0.195\linewidth]{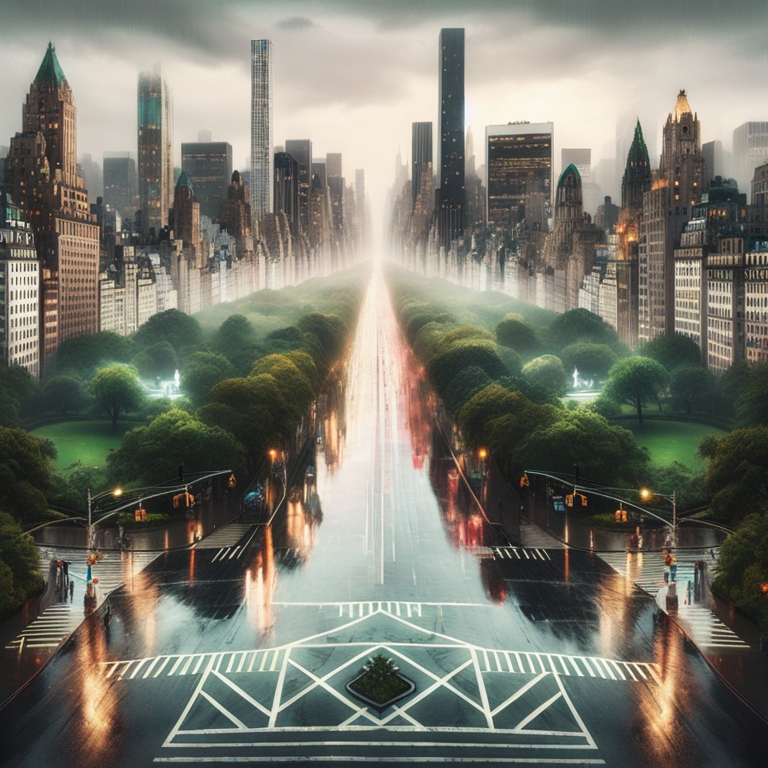}
        \vspace{-0.2em}\caption{Get rid of the traffic.}
    \end{subfigure}\\[0.1em]
    \begin{subfigure}[b]{\textwidth}
        \centering
        \includegraphics[width=0.195\linewidth]{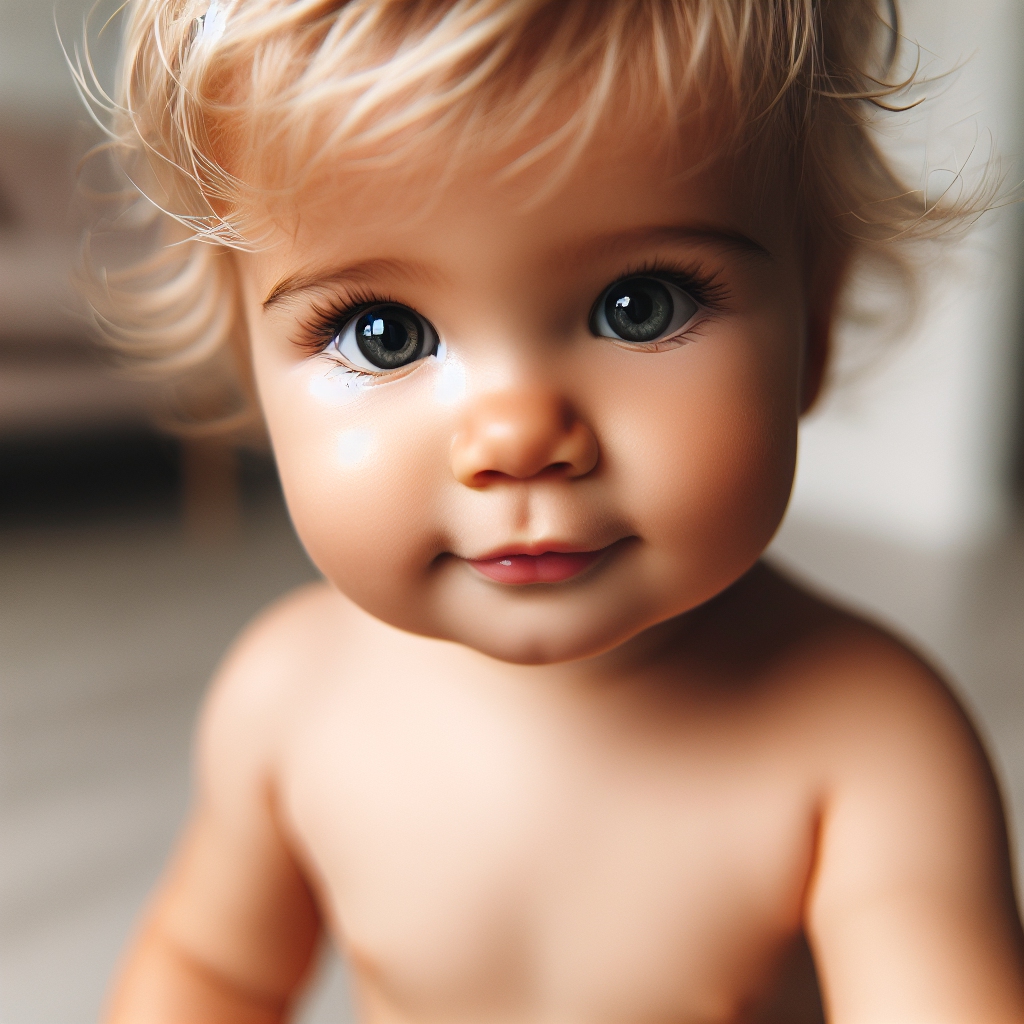}\hfill
        \includegraphics[width=0.195\linewidth]{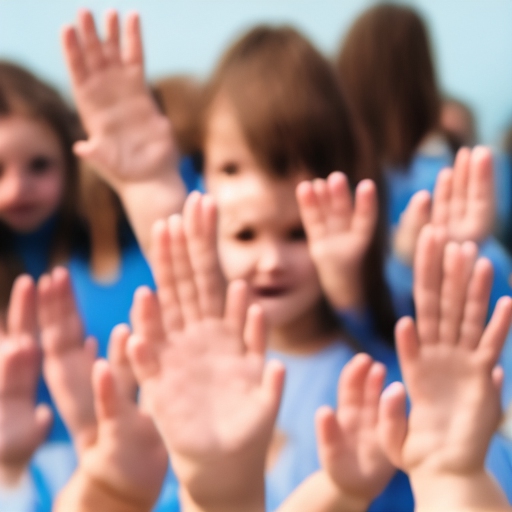}\hfill
        \includegraphics[width=0.195\linewidth]{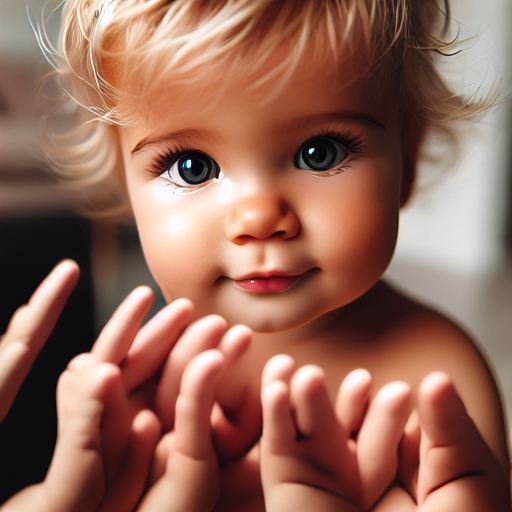}\hfill
        \includegraphics[width=0.195\linewidth]{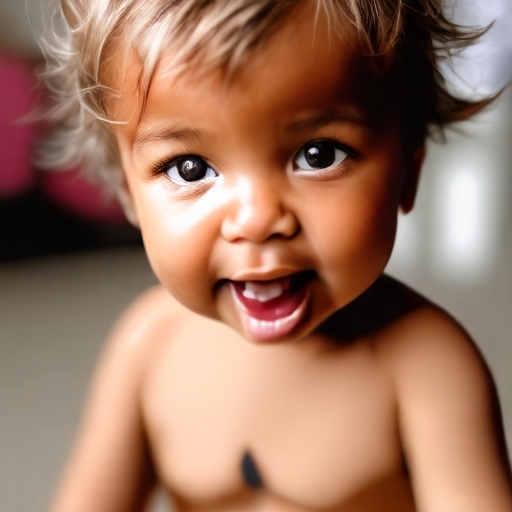}\hfill
        \includegraphics[width=0.195\linewidth]{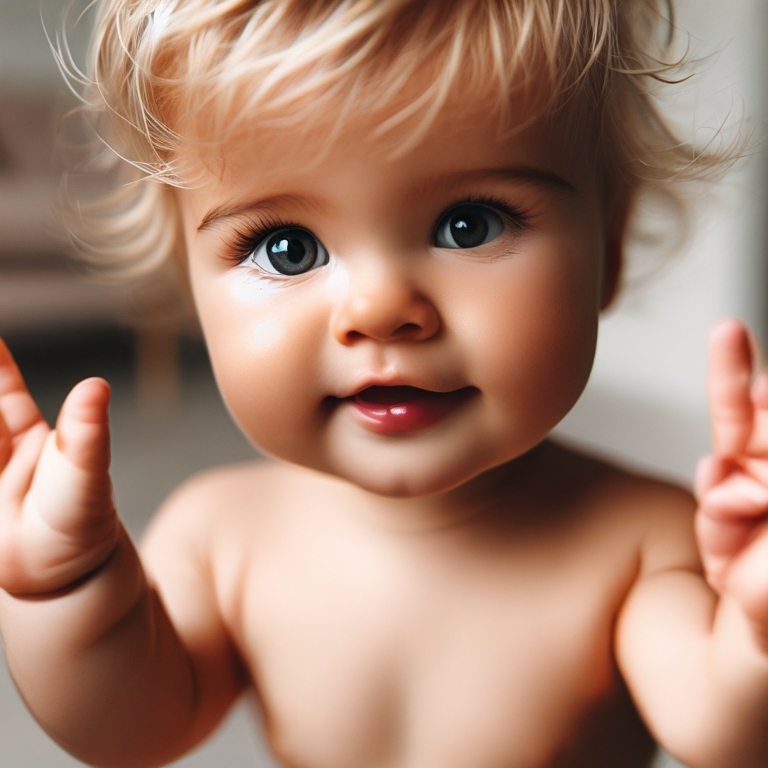}
        \vspace{-0.2em}\caption{Let the subject raise up hands.}
    \end{subfigure}\hfill
    \vspace{-0.6em}
    \caption{Example Results of different methods on the HQ-Edit benchmark.}\vspace{-0.5em}
    \label{fig:baseline_hqedit}
\end{figure}

Fig~\ref{fig:baseline_hqedit} shows some qualitative examples of our method and baselines on the HQ-Edit benchmark. A major difference between our method is that our method could understand rather ambiguous instructions and when performing fine-grained editing with a higher success rate.

\begin{figure}[t]
\includegraphics[width=1.0\linewidth]{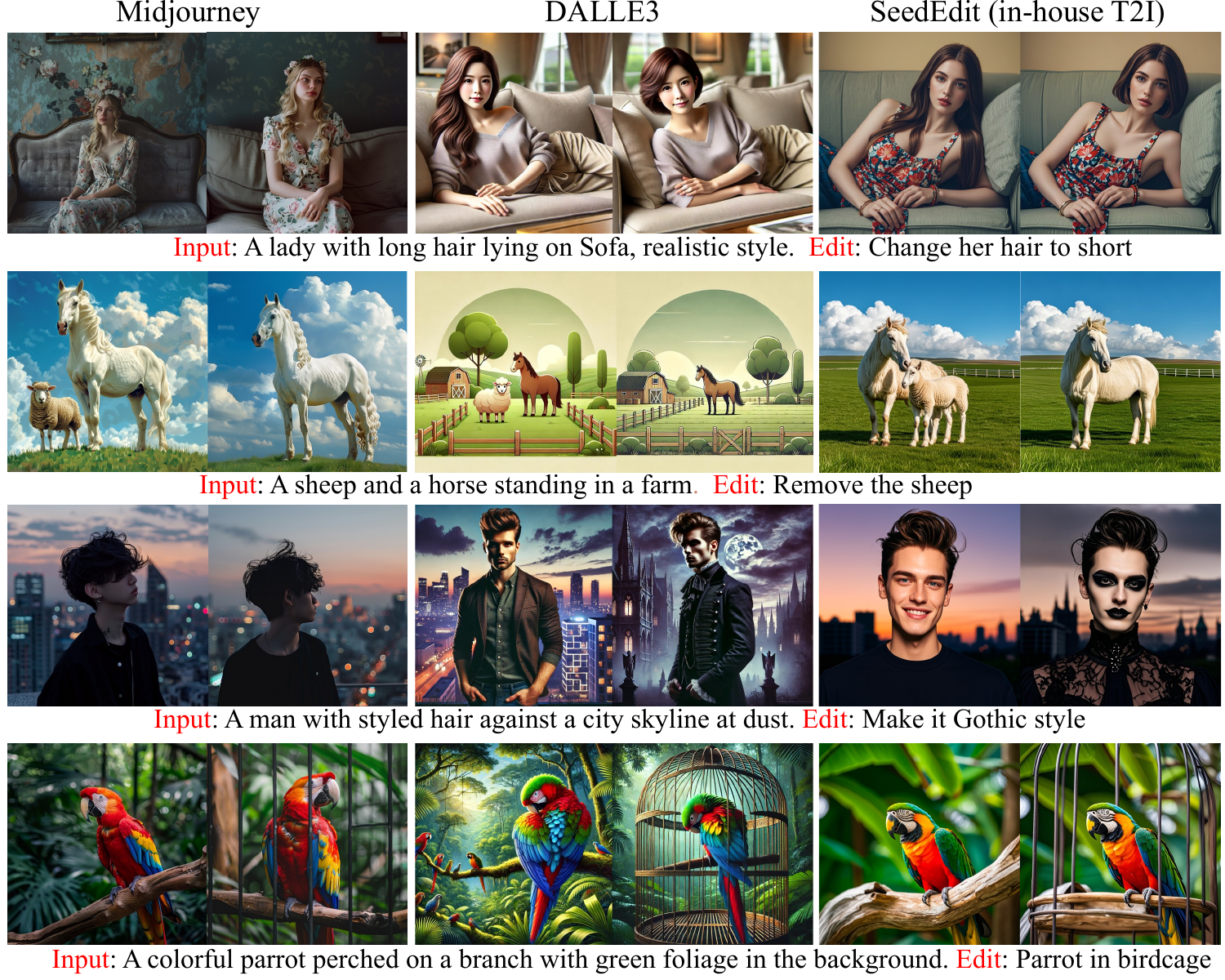}
\caption{Example results from different products for editing based image generation.}\vspace{-0.5em}
\label{fig:prod_compare}
\end{figure}

Lastly, we compare the image editing capabilities of SeedEdit (in-house T2I model) with other commercial SoTA tools, such as DALLE3 Edit\footnote{https://openai.com/index/hello-gpt-4o/} and Midjourney\footnote{https://docs.midjourney.com/docs/the-web-editor}, which allow the editing of self-generated images. Fig.~\ref{fig:prod_compare} presents a qualitative comparison of the results. In general, both DALLE3 and Midjourney tend to introduce more unintended content changes beyond the specified editing prompt. Between the two, Midjourney produces more aesthetically pleasing images, while DALLE3 demonstrates superior adherence to the prompt instructions. In contrast, as shown in the last column, SeedEdit strikes a better balance, offering more precise editing that closely follows the given instructions. Furthermore, we conducted an internal user study that indicated a strong preference for the results generated by our method.

\begin{table}[t]
\hspace{-0.5em}
\small
\captionsetup{font=small}
\setlength{\tabcolsep}{8pt}
\renewcommand*{\arraystretch}{1.15}
\begin{center}
\begin{tabularx}{1.0\linewidth}{X ccc c ccc}
\toprule
& \multicolumn{3}{c}{\textbf{HQ-Edit}} &\quad& \multicolumn{3}{c}{\textbf{Emu Edit}} \\\cline{2-4}\cline{6-8}
\renewcommand*{\arraystretch}{1.0}
Model & GPT$\uparrow$ & CLIP\textsubscript{dir}$\uparrow$ & CLIP\textsubscript{img}$\uparrow$ && GPT$\uparrow$ & CLIP\textsubscript{dir}$\uparrow$ & CLIP\textsubscript{img}$\uparrow$ \\
\midrule
Prompt-to-Prompt & $26.93$ & $0.0811$ & $0.7462$ && $12.69$ & $0.0488$ & $0.6568$ \\
Instruct-Pix2Pix & $47.50$ & $0.1224$ & $0.8390$ && $31.39$ & $0.0726$ & $0.8092$ \\
MagicBrush & $47.51$ & $0.1287$ & $0.8008$ && $44.25$ & $0.0856$ & $0.7930$ \\
Emu Edit & N/A & N/A & N/A && $64.51$ & $0.1094$ & $\mathbf{0.8206}$ \\
UltraEdit & $54.17$ & $0.1473$ & $0.8281$ && $46.95$ & $0.0933$ & $0.8072$ \\\hline
SeedEdit (SDXL)     & $71.24$ & $0.1656$ & $\mathbf{0.8698}$  && $66.48$ & $\mathbf{0.1162}$ & $0.8025$ \\
SeedEdit (in-house T2I)    & $\mathbf{78.54}$ & $\mathbf{0.1766}$ & $0.8524$      && $\mathbf{75.03}$ & $0.1137$ & $0.7875$ \\
\bottomrule
\end{tabularx}
\vspace{-0.5em}\caption{Quantitative evaluation on image editing benchmarks.}
\label{tab:compare_baseline}
\end{center}
\end{table}
\begin{figure}[t]
    \newcolumntype{Y}{>{\centering\arraybackslash}X}
    \begin{tabularx}{\linewidth}{YYYYYY}
    \small Input & \small  Instruct-P2P & \small MagicBrush & \small UltraEdit & \small Emu Edit & \small  SeedEdit (SDXL) \\
    \end{tabularx}
    \begin{subfigure}[b]{\textwidth}
        \centering
        \includegraphics[width=0.166\linewidth]{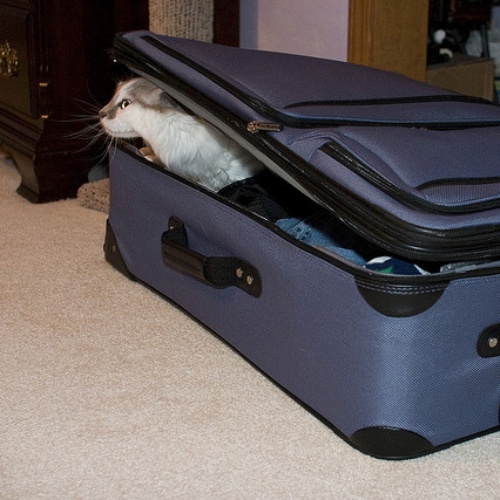}\hfill
        \includegraphics[width=0.166\linewidth]{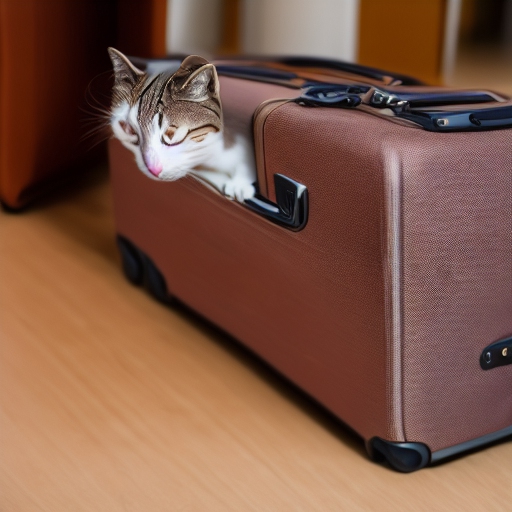}\hfill
        \includegraphics[width=0.166\linewidth]{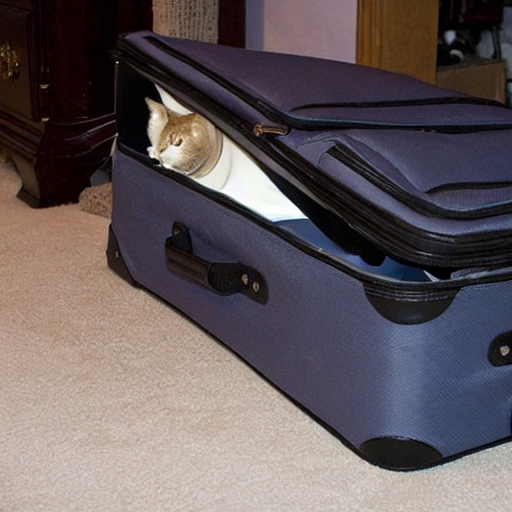}\hfill
        \includegraphics[width=0.166\linewidth]{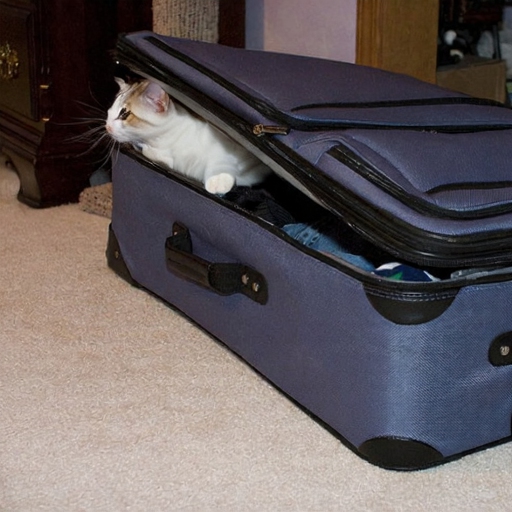}\hfill
        \includegraphics[width=0.166\linewidth]{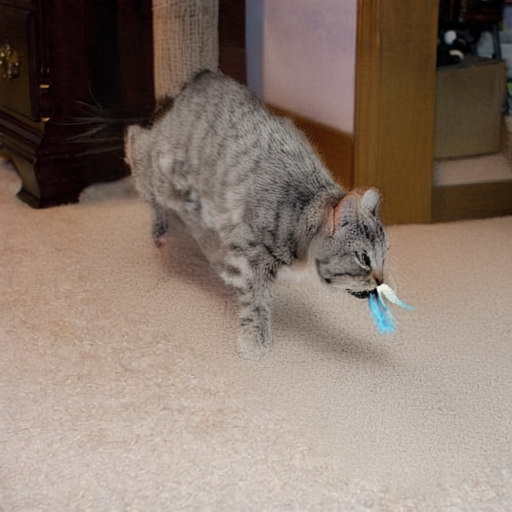}\hfill
        \includegraphics[width=0.166\linewidth]{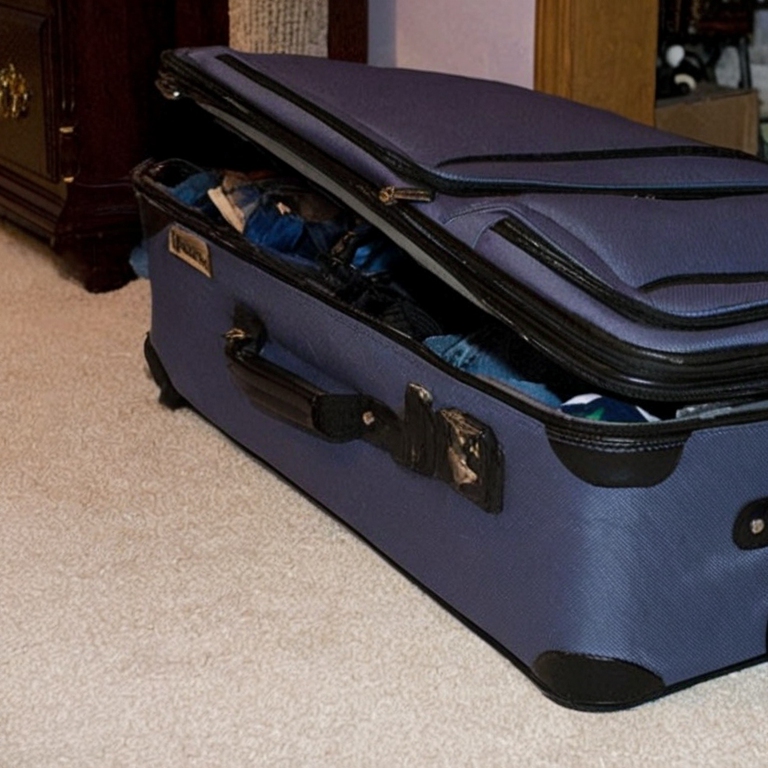}\hfill
        \vspace{-0.2em}\caption{Get rid of the cat peering out from the suitcase.}
    \end{subfigure}\\[0.1em]
    \begin{subfigure}[b]{\textwidth}
        \centering
        \includegraphics[width=0.166\linewidth]{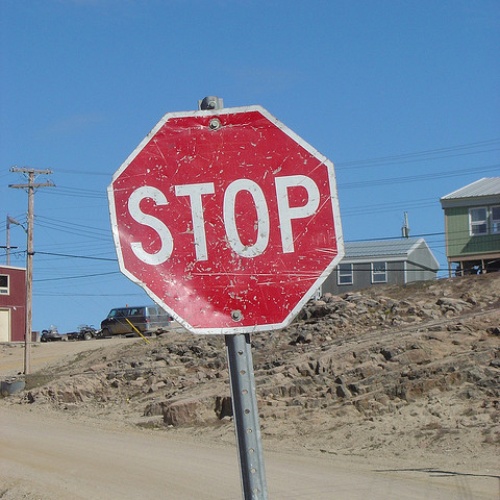}\hfill
        \includegraphics[width=0.166\linewidth]{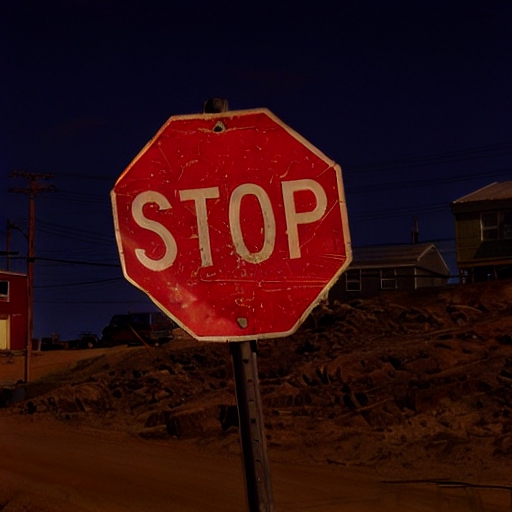}\hfill
        \includegraphics[width=0.166\linewidth]{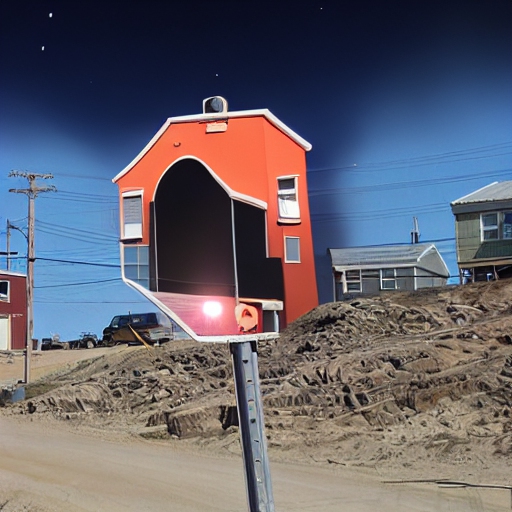}\hfill
        \includegraphics[width=0.166\linewidth]{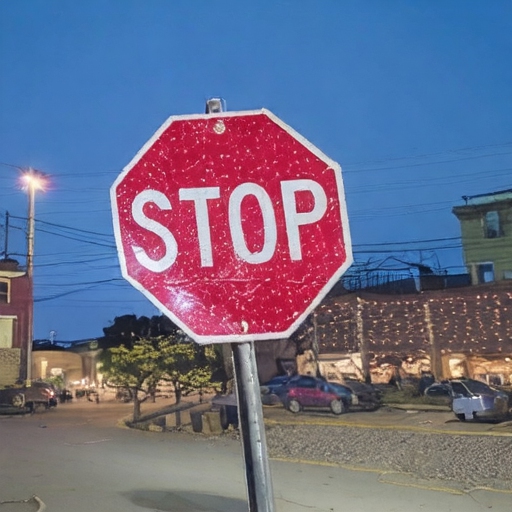}\hfill
        \includegraphics[width=0.166\linewidth]{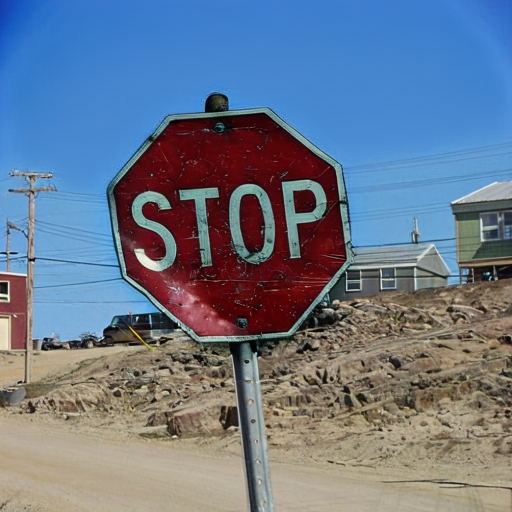}\hfill
        \includegraphics[width=0.166\linewidth]{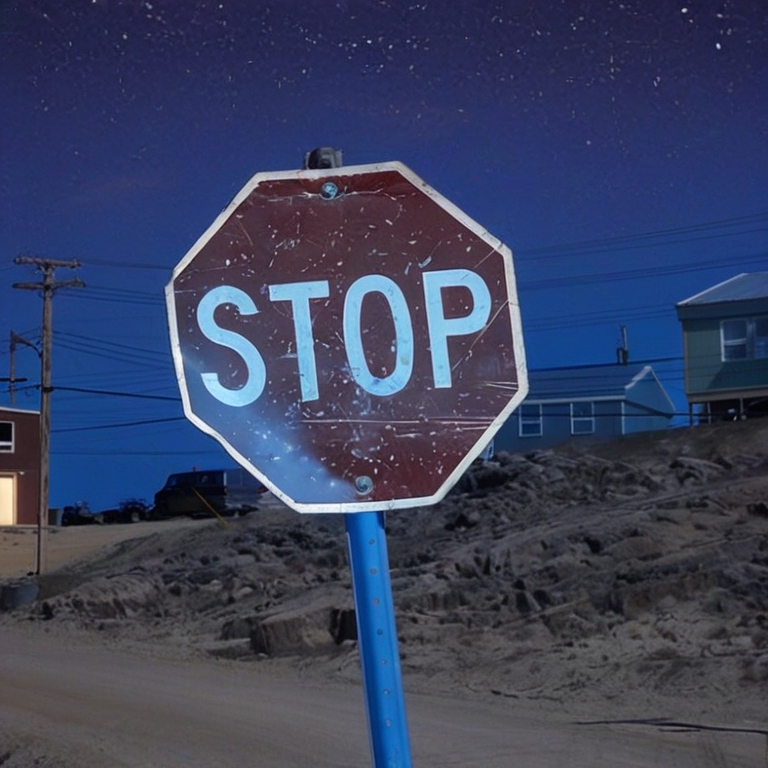}\hfill
        \vspace{-0.2em}\caption{Change the time of the day to night}
    \end{subfigure}\hfill
    \vspace{-0.6em}
    \caption{Example results of different methods on the Emu Edit benchmark.}
    \label{fig:baseline_emu}
\end{figure}


\section{Conclusion}
In this work, we introduced SeedEdit, a progressive alignment framework to adapt a pre-trained T2I diffusion model to image editing model, which maximizes both prompt alignment and image consistency. An causal diffusion model is proposed to take both images and texts as conditions for image generation. An iterative data generation and fine-tuning framework is proposed to align the diffusion towards precise image editing. Experimental results demonstrate that our method yields superior results compared to existing methods by a large margin.

\newpage
\bibliography{iclr2024_conference}
\bibliographystyle{iclr2024_conference}

\appendix

\end{document}